\documentclass{article}

\usepackage{microtype}
\usepackage{graphicx}
\usepackage{subcaption}
\usepackage{booktabs} % for professional tables

\usepackage{hyperref}

\usepackage[preprint]{icml2026}
\usepackage{amsmath}
\usepackage{amssymb}
\usepackage{mathtools}
\usepackage{amsthm}
\usepackage{tabularx}
\usepackage[table]{xcolor}

\usepackage{makecell}  % 解决 \makecell 报错
\usepackage{colortbl}  % 解决 \rowcolor, \cellcolor 报错
\usepackage{array}     % 解决 !{\vrule ...} 竖线报错
\usepackage{booktabs}  % 虽然你这里没用 \toprule，但通常建议留着
\usepackage{xcolor}    % 颜色支持
\usepackage{graphicx}  % 解决 \resizebox, \rotatebox (icml可能已加载，但重复写无害)
\usepackage{multirow}  % 如果之后用到多行合并
\usepackage{xspace}
\usepackage{multirow}
\usepackage{caption}
\definecolor{teal}{RGB}{0,128,128} % 自定义颜色，防止报错
\definecolor{yellow}{RGB}{255,215,0}

\newcommand{\eg}{\textit{e.g.}\xspace}
\newcommand{\ie}{\textit{i.e.}\xspace}

\definecolor{DeepBlue}{HTML}{156082}
\definecolor{LightBlue}{HTML}{E6F0FF} % 近似 blue!5 但更可控

\usepackage[capitalize,noabbrev]{cleveref}

\theoremstyle{plain}

\theoremstyle{definition}

\theoremstyle{remark}

\usepackage[textsize=tiny]{todonotes}

\icmltitlerunning{GeoSense: Internalizing Geometric Necessity Perception for Multimodal Reasoning}

\begin{document}

\twocolumn[
  \icmltitle{GeoSense: Internalizing Geometric Necessity Perception \\
  for Multimodal Reasoning}



  \icmlsetsymbol{equal}{*}

% Ruiheng Liu, Haihong Hao, Mingfei Han, Xin Gu, Kecheng Zhang, Changlin Li, Xiaojun Chang

  \begin{icmlauthorlist}
    \icmlauthor{Ruiheng Liu}{equal,ustc}
    \icmlauthor{Haihong Hao}{equal,ustc}
    \icmlauthor{Mingfei Han}{equal,ustc,mbzuai}
    \icmlauthor{Xin Gu}{ucas}
    \icmlauthor{Kecheng Zhang}{ustc}
    \icmlauthor{Changlin Li}{stanford}
    \icmlauthor{Xiaojun Chang}{ustc,mbzuai}
      \url{https://water-wood-rain.github.io/Geosense/}
  \end{icmlauthorlist}

  \icmlaffiliation{ustc}{University of Science and Technology of China, Anhui, China}
  \icmlaffiliation{mbzuai}{Mohamed Bin Zayed University of Artificial Intelligence, Abu Dhabi, UAE}
  \icmlaffiliation{ucas}{University of Chinese Academy of Sciences, Beijing, China}
  \icmlaffiliation{stanford}{Stanford University, CA, USA}

  \icmlcorrespondingauthor{Xiaojun Chang}{xjchang@ustc.edu.cn}
  % \icmlcorrespondingauthor{Mingfei Han}{hmf282@gmail.com}

  \icmlkeywords{Machine Learning, ICML}

  \vskip 0.3in
]



\printAffiliationsAndNotice{\icmlEqualContribution}

\begin{abstract}
  Advancing towards artificial superintelligence requires rich and intelligent perceptual capabilities. A critical frontier in this pursuit is overcoming the limited spatial understanding of Multimodal Large Language Models (MLLMs), where geometry information is essential. Existing methods often address this by rigidly injecting geometric signals into every input, while ignoring their necessity and adding computation overhead.
  Contrary to this paradigm, our framework endows the model with an awareness of perceptual insufficiency, empowering it to autonomously engage geometric features in reasoning when 2D cues are deemed insufficient.
  To achieve this, we first introduce an independent geometry input channel to the model architecture and conduct alignment training, enabling the effective utilization of geometric features. Subsequently, to endow the model with perceptual awareness, we curate a dedicated spatial-aware supervised fine-tuning dataset. This serves to activate the model’s latent internal cues, empowering it to autonomously determine the necessity of geometric information.
  Experiments across multiple spatial reasoning benchmarks validate this approach, demonstrating significant spatial gains without compromising 2D visual reasoning capabilities, offering a path toward more robust, efficient and self-aware multi-modal intelligence.
  % This document provides a basic paper template and submission guidelines.
  % Abstracts must be a single paragraph, ideally between 4--6 sentences long.
  % Gross violations will trigger corrections at the camera-ready phase.
\end{abstract}
   
\section{Introduction}
\label{sec:intro}

\begin{figure}[t!]
    \centering
    \includegraphics[width=1\linewidth]{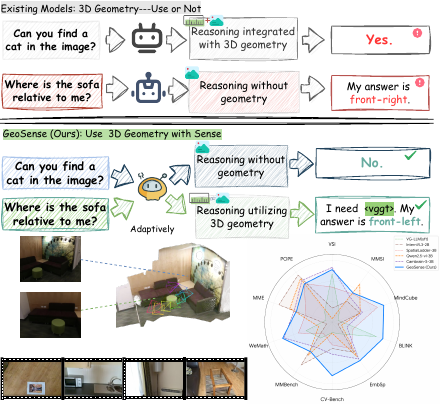}
    \caption{\textbf{Adaptive Geometric Reasoning with GeoSense.} (Top) Existing MLLMs typically adopt a static approach to 3D geometry, either ignoring it or rigidly fusing it, which leads to confusion in general tasks or failures in spatial reasoning. (Bottom) GeoSense introduces an adaptive mechanism (\textit{Use with Sense}) that requests geometric features only when necessary. As shown in the radar chart, this flexibility allows GeoSense to achieve SOTA performance across both general visual benchmarks (\eg, MMBench~\cite{liu2024mmbenchmultimodalmodelallaround}, WeMath~\cite{qiao2024wemathdoeslargemultimodal}) and spatial reasoning tasks (\eg, VSI-Bench~\cite{yang2025thinking} and MindCube~\cite{yin2025spatialmentalmodelinglimited}).}
    \label{fig:example}
    \vspace{-4mm}
\end{figure}

% The development of multimodal large language models (MLLMs) has enabled new intelligent solutions in domains such as autonomous driving, embodied AI, and industrial robotics~\cite{xie2025s4,li2024manipllm,szot2025multimodal,ji2025robobrain}. However, as these models are increasingly deployed in real-world settings, spatial reasoning has emerged as a critical bottleneck that affects both reliability and decision accuracy. Previous studies indicate that improving spatial reasoning depends on the effective perception of multiple spatial cues in the physical world~\cite{yang2025thinking,liu2025spatialssrlenhancingspatialunderstanding,ning2025enhancing,daxberger2025mm,wu2025spatial,zheng2025learning}, and the incorporation of 3D information has been shown to significantly enhance spatial understanding in MLLMs, which has led to a growing body of work leveraging 3D cues to assist vision-language action models~\cite{zitkovich2023rt,qu2025spatialvla,kim2024openvla}, embodied navigation agents~\cite{zhou2024navgpt,zheng2024towards,zhang2024navid}, and video-based 3D knowledge learning systems~\cite{zhang2024navid,qi2025gpt4scene,zheng2025video}.

Multimodal Large Language Models (MLLMs) have revolutionized domains such as autonomous driving, embodied AI, and industrial robotics~\cite{xie2025s4,li2024manipllm,szot2025multimodal,ji2025robobrain}. However, as these models transition to real-world deployment, spatial reasoning has emerged as a critical bottleneck that affects both reliability and decision accuracy. While recent works ~\cite{yang2025thinking,liu2025spatialssrlenhancingspatialunderstanding,daxberger2025mm,wu2025spatial,zheng2025learning} have shown that incorporating geometric information significantly enhances spatial understanding, the effective integration of these cues remains a challenge.

Due to the scarcity of native 3D data such as depth maps and point clouds~\cite{wu2025spatial}, research has shifted toward reconstructing 3D representations from 2D visual inputs using foundation models like VGGT~\cite{wang2025vggt} and MoGe-2~\cite{wang2025moge}). 
Enabled by these 3D foundation models, recent approaches have begun injecting geometry-encoded embeddings derived from 2D visual inputs directly into the reasoning pipeline~\cite{zheng2025learning,wu2025spatial}. 
For example, Spatial-MLLM~\cite{wu2025spatial} projects the geometric features from VGGT~\cite{wang2025vggt} into the pretrained vision-language space to provide 3D cues during inference, fostering an implicit internalized 3D awareness.
However, these approaches typically follow a tightly coupled design that treats 3D geometric information as a mandatory input, regardless of the specific task requirements. 
This assumes geometry as a universal necessity, yet in practice, its utility is highly context-dependent.
For instance, while depth cues are vital for spatial navigation, they serve as irrelevant noise for non-spatial tasks such as table OCR or plane geometry. This forced activation of geometric signals can degrade general reasoning capabilities by introducing unnecessary complexity and degrades general reasoning performance.

We identify this limitation as the perception information gap: the discrepancy between the information a model perceives from standard 2D inputs, and the specific context and domain knowledge required to solve complex spatial reasoning problems. 
Current MLLMs lack the internal awareness to recognize this gap or assess their own information requirements. To address this, we introduce \textbf{\textit{GeoSense}}, a framework that internalizes the perception of geometric necessity. Our approach endows the model with the intrinsic capability to autonomously decide whether to integrate 3D geometric cues based solely on the input content and query, without requiring explicit spatial instructions.

% Specifically, we design a training framework to optimize Multimodal Large Language Models (MLLMs) equipped with an independent 3D feature input path. We treat a 3D foundation model as the 3D encoder and feed its output features into the model as independent geometric embeddings. In the first stage, a dedicated projection layer maps the 3D features and 2D visual embeddings into a unified embedding space. We then apply supervised fine-tuning to align the projected 2D and 3D features with text embeddings, enabling the model to effectively utilize independent 3D cues for reasoning.
Specifically, our framework training consists of two stages. First is the geometry alignment. 
% We implement this via an independent geometry input channel, allowing geometric features to serve as an on-demand resource rather than a mandatory burden.
We treat geometric information as a standalone modality via an independent geometry input channel rather than element-wise addition~\cite{zheng2025learning,wu2025spatial} to 2D features. This allows geometric features to serve as an on-demand resource rather than a mandatory burden.
The second is spatial perception tuning. We curate a model-adaptive, spatially-aware dataset by evaluating the model’s own performance discrepancies with and without geometric features, and reformulate these variances into training signals. This allows the model to learn its own empirical priors, superseding rigid, human-defined rules.

As a result, our model dynamically modulates its reliance on geometry input, achieving superior spatial reasoning while preserving general-purpose visual intelligence. Releasing dynamically the burden of an additional spatial encoder, our approach is particularly suitable for small scale MLLMs deployed on edge devices.
In summary, our contributions are summarized as follows.
\begin{itemize}
    % Point 1: 整体框架
    \item We propose a two-stage training framework that enables the model to autonomously select reasoning pathway based on context-specific information demands.
    % , rather than always relying on the 3D input channel.
    % Point 2: 核心机制（即你刚刚翻译的这点，强调数据驱动替代规则）
    % \item We construct a training dataset encapsulating the model's intrinsic empirical priors to supersede human-defined rule systems, thereby achieving the internalization of task-aware spatial necessity.
    \item We construct a model-adaptive data curation pipeline that captures intrinsic empirical priors of MLLMs, internalizing the perception of geometry necessity without relying on hand-crafted rule systems.
    % Point 3: 实验结果
    % \item We demonstrate that our method improves spatial reasoning performance while preserving the model's general-purpose reasoning ability.
    \item We show that our method significantly improves spatial reasoning performance while maintaining robust general-purpose reasoning across diverse benchmarks.
\end{itemize}
% \begin{itemize}
%     \item We propose a two-stage training framework that enables the model to autonomously select its reasoning pathway based on task-specific information demands, rather than always relying on the 3D input channel.
%     \item we
%     \item We demonstrate that our method improves spatial reasoning performance while preserving the model's general-purpose reasoning ability.
% \end{itemize}
% In the second stage, we perform RL-based cold-start training on a curated dataset of approximately 4K samples designed to teach the model when 3D information is required, based on task-dependent information gaps. In the final RL stage, the model is further trained on a collection of visually grounded reasoning tasks, refining its ability to infer information needs and selectively activate the 3D pathway during inference \textbf{(reward design details TBD)}. Experimental results \textbf{(TBD)} show that our method achieves strong spatial reasoning performance while preserving the general reasoning capabilities acquired during pretraining. In summary, the main contributions of this work are as follows.

\section{Related Work}
\label{sec:related}

\subsection{MLLMs for Visual Scene Understanding}
\label{subsec:mllm4vision}
The evolution of Multimodal Large Language Models (MLLMs) has significantly advanced visual scene understanding. Representative approaches, such as BLIP-2~\cite{li2023blip} and LLaVA~\cite{liu2023visual,liu2024improved}, align visual encoders with LLMs via lightweight projectors, enabling robust 2D perception. Recent state-of-the-art models like Qwen2.5-VL~\cite{bai2025qwen2} and InternVL-3~\cite{zhu2025internvl3} further push performance boundaries through high-resolution inputs and dynamic token allocation. However, these 2D-native architectures inherently lack explicit geometric representations, leading to systematic failures in tasks requiring structural consistency and perspective manipulation~\cite{Zhu_2024_CVPR,liu2025coarse}. 

To bridge this gap, research has branched into 2.5D strategies injecting depth cues~\cite{zhu2025llava,huang2025mllm} and 3D-native models processing point clouds~\cite{xu2024pointllm,mao2025spatiallm}. While effective, 3D-native methods often entangle appearance and geometry at the input level. More recently, hybrid approaches have attempted to fuse features from 3D foundation models directly into 2D visual embeddings~\cite{wu2025spatial,zheng2025learning,chen2025think}. Unlike these fusion-based methods, which risk compromising the model's original planar understanding, our approach treats 3D features as a distinct, independent input modality, ensuring that 2D visual reasoning remains unaffected.

\subsection{Visual Spatial Intelligence}
\label{subsec:spatial-intel}
Spatial intelligence encompasses relational reasoning (\eg, depth, topology) and viewpoint transformation~\cite{yang2025thinking,song2025robospatial}. Despite excelling in general visual tasks, MLLMs often struggle with these spatial dimensions due to the scarcity of 3D annotated data~\cite{hudson2019gqa,li2024topviewrs}. Early solutions relied on synthetic datasets to teach basic spatial concepts~\cite{chen2024spatialvlm,cheng2024spatialrgpt}, but these failed to capture dynamic real-world complexities. 

With the advent of powerful geometric estimation models~\cite{wang2025vggt,wang2025moge}, recent works utilize predicted 3D priors to enhance MLLMs without costly 3D annotations. Systems like SpatialMLLM~\cite{wu2025spatial} and VGLLM~\cite{zheng2025learning} integrate these priors into the visual latent space. However, such indiscriminate incorporation forces the model to process 3D information even when unnecessary, which has been shown to impair performance on general reasoning tasks like mathematics or OCR~\cite{foroutan2025wikimixqa,wang2025mathcoder}. In contrast, our work introduces an adaptive mechanism that selectively incorporates 3D information based on task necessity, thereby enhancing spatial understanding while preserving the model's general reasoning capabilities acquired during pretraining.
\section{Method}
% Our method is aimed to enabling the multimodal model to detect 2D perception insufficiency and autonomously utilize 3D geometric cues. Since pilot experiments reveal that indiscriminate 3D incorporation can be suboptimal, we propose a decoupled architecture where 3D features serve as independent input channels for selective usage. Validated on multiple benchmarks, a two-stage supervised fine-tuning strategy effectively instills this adaptive capability in the model.

We propose \textbf{\textit{GeoSense}}, a multimodal framework designed to adaptively integrate 3D cues based on its internal geometric necessity perception. Unlike conventional approaches that rigidly fuse geometric features, GeoSense employs a decoupled framework where geometric information serves as an on-demand resource. This capability is realized through a novel request mechanism learned via a two-stage supervised fine-tuning strategy, allowing the model to autonomously determine when geometric injection is strictly necessary.

\begin{figure*}[t!] % 关键点1：加星号 *；关键点2：通常使用 [t] 置顶
    % \vskip -0.1in % ICML 模版建议的顶部间距
    % \vspace{-0.2mm}
    \begin{center}
        % 关键点3：宽度通常设为 \linewidth 或 \textwidth (占满通栏)
        % 如果觉得太大，可以改为 0.8\linewidth
        \centerline{\includegraphics[width=0.95\linewidth]{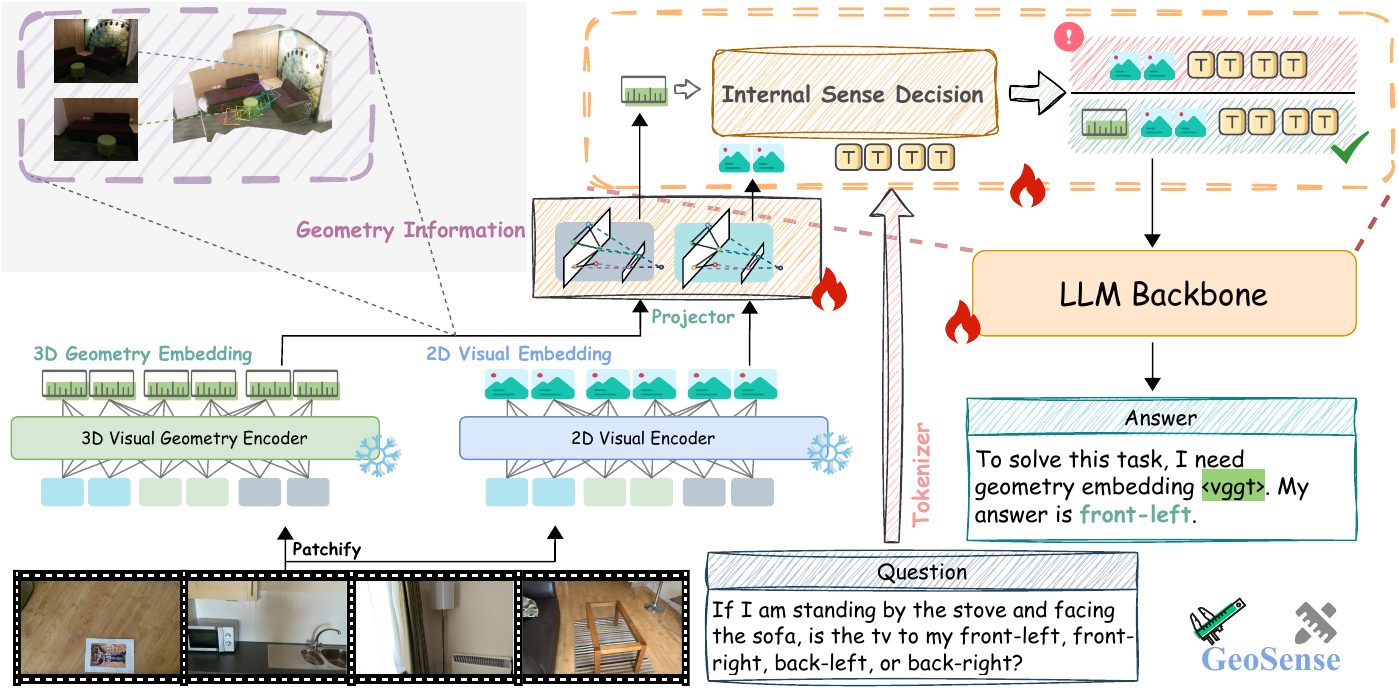}}
        
        \caption{\textbf{Architectural Overview of GeoSense.} We integrate a 3D visual geometry encoder alongside a standard 2D visual encoder, both of which are kept frozen to preserve pretrained representations. Dedicated projection layers map these features into a unified embedding space for the LLM backbone. During inference, the model dynamically makes an ``Internal Sense Decision'' based on the 2D visual and textual prompt. If the latent state triggers a geometry request (\eg, via the \texttt{<vggt>} token), 3D embeddings are concatenated to the sequence for a second re-inference pass. During training, only the projection layers and the LLM backbone are optimized.}
        \label{fig:pipeline}
    \end{center}
    % \vskip -0.2in % ICML 模版建议的底部间距修正
    \vspace{-3.5mm}
\end{figure*}

\subsection{Geometry Cues Are Not Universally Beneficial}
\label{sec:pilot}

% To realize our objective, endowing the model with the capacity for spatial perception, appropriate training data is indispensable, complementing our architectural separation of input channels. We initially attempted to manually codify a set of criteria to instruct the model on when to employ 3D information. However, upon reflection, we concluded that a universal ``Gold Standard'' covering all scenarios is elusive. For instance, a photorealistic image might merely require object counting (\eg, kittens) without depth reasoning, whereas a 2D line drawing could represent a complex solid geometry problem requiring distinct 3D cues. Consequently, the boundary of necessity is blurred; spatial awareness in humans is not governed by rigid rules but is rather rooted in empirical experience derived from lifelong interaction with the physical world.

% To equip the model with adaptive spatial perception, we must first determine \textit{when} geometry is actually needed. 
% % Initially, we attempted to manually codify a set of rules for 3D feature usage. 
% However, defining a universal heuristic proved intractable. For example, a photorealistic image might require simple object counting (solvable with 2D cues), whereas a simple line drawing could present a complex solid geometry problem requiring depth information. 
% Thus, the boundary of necessity is blurred; much like human spatial awareness, which relies on implicit intuition rather than rigid rules, 
% the need for geometry is highly context-dependent.

To realize adaptive spatial perception, we must first determine \textit{when} geometry is actually needed. However, defining a universal heuristic for this is intractable. While a photorealistic image may only require 2D object counting, a simple line drawing might demand complex 3D depth reasoning. Much like human spatial awareness, the necessity of geometry is highly context-dependent and rooted in implicit intuition rather than rigid rules.

% Drawing inspiration from this, we hypothesize that the model might also have acquired ineffable empirical knowledge during its pre-training and fine-tuning phases. Specifically, if we cannot prescribe a priori rules, why not empower the model to articulate its own needs? To this end, we conducted a pilot study using a VG-LLM that had been well-trained on spatial data with aligned 3D features. We fed a mixture of spatial reasoning and general visual question-answering (VQA) datasets into the model and performed inference under two distinct conditions: utilizing explicit 3D inputs and suppressing them (via zero-padding).

% Hypothesizing that the model could develop similar latent priors during pre-training, we aimed to empower it to articulate its own needs rather than relying on hand-crafted rules. 
% To validate this, we conducted a pilot study using a standard VG-LLM~\cite{} trained on spatial data (\eg, multi-view captures of a scene). We evaluated the model on a mixture of spatial reasoning~\cite{} and general VQA datasets~\cite{} under two conditions: \textit{i.} with geometry embedding injected, and \textit{ii.} with geometry inputs suppressed (via zero-padding).

Hypothesizing that MLLMs can develop similar latent priors, we aimed to empower it to articulate its own needs rather than relying on hand-crafted rules. To validate this, we evaluated a standard VG-LLM~\cite{zheng2025learning} on a mixture of spatial reasoning~\cite{wu2025spatial} and general VQA dataset~\cite{zhang2024direct} under two conditions: (\textit{i}) with geometry embeddings injected, and (\textit{ii}) with geometry inputs suppressed via zero-padding.
% Analysis of approximately 700K samples revealed a compelling distribution. For $\sim$67\% of the data, the model answered correctly regardless of 3D feature usage, while for $\sim$25\%, it failed under both conditions; we attribute these outcomes primarily to the limits of the model's base capacity. The remaining $\sim$8\% constitutes the critical signal: 5\% of the samples required 3D geometric features for a correct response, whereas, counter-intuitively, the remaining 3\% yielded better results when 3D features were omitted. Notably, even within datasets specifically curated for spatial reasoning, approximately 4\% of samples exhibited performance degradation upon the introduction of 3D geometric information.
Quantitative analysis of approximately 700k samples revealed a critical insight. For $\sim$67\% of the samples, the model was robust regardless of geometry usage, while $\sim$25\% failed under both conditions (likely due to base capacity limits). The remaining $\sim$8\% highlighted the ``double-edged sword'' nature of geometric injection: while 5\% of samples \textit{required} 3D features for a correct response, the remaining 3\% actually suffered performance degradation when geometric features were added. 

This counter-intuitive finding, that geometry can act as noise, was consistent across both general benchmarks (where geometry suppression improved performance) and spatial benchmarks (where geometry inclusion was vital), comparing between VG-LLM~\cite{zheng2025learning} and Qwen2.5-VL-3B~\cite{bai2025qwen2}, as shown in Table~\ref{tab:main_results}.
% These empirical findings confirm that rigid integration is suboptimal and establish the empirical foundation for our adaptive data construction.
\textit{Crucially, this issue is exacerbated by scale.} Enlarging the training scale (from 385k to 940k) and increasing spatial data under a rigid integration paradigm (\ie, VG-LLM) further degrades the performance on general visual benchmarks, leading to severe hallucinations (POPE~\cite{li2023evaluatingobjecthallucinationlarge} drops from 86.9 to 74.2) and cross-modal confusion (average score drops from 52.0 to 48.2). 
These findings confirm that rigid integration is suboptimal, establishing the empirical foundation for our adaptive data construction and necessitating the adaptive framework with our GeoSense.
\subsection{Independent Geometry Adaptation}
\label{sec:arch}

To achieve autonomous spatial awareness without compromising native 2D perception, it requires a decoupled input architecture. Unlike prior approaches that fuse geometric features via element-wise summation~\cite{zheng2025learning,wu2025spatial}, we treat 3D geometry as an independent, on-demand modality.
This design ensures that the high-resolution 2D visual stream, responsible for general recognition, remains unpolluted by geometric embeddings.
% unless explicitly invoked by the model's internal decision mechanism. By isolating these features into distinct channels, we preserve the LLM's base reasoning capacity while providing a dedicated pathway for precise spatial information.

\paragraph{Model Architecture.} We utilize the visual encoder of Qwen2.5-VL-3B~\cite{bai2025qwen2} and VGGT~\cite{wang2025vggt} as our 2D and geometric encoders. Upon receiving visual input, both encoders extract features which are projected via dedicated MLP modules to a shared embedding space. These projected features are then populated into specific placeholders, \texttt{<|vision\_pad|>} and \texttt{<|vggt\_pad|>}, to construct a comprehensive multimodal prompt containing discrete 2D, geometric, and textual components.

\paragraph{Geometry sequence formulation.}
Formally, given a visual input \(\mathcal{V}\), the 3D encoder \(\mathcal{E}_{\text{geo}}\) extracts structured geometric features:
\begin{equation}
    F_{\text{geo}} = \mathcal{E}_{\text{geo}}(\mathcal{V}) \in \mathbb{R}^{T \times D_{\text{3D}}},
    \label{eq:geom-feat}
\end{equation}
where \(T\) denotes the sequence length and \(D_{\text{3D}}\) the raw feature dimension. To align these with the MLLM's embedding space (\(D_{\text{MM}}\)), we apply a learnable linear projection \(W^{3D}_{\text{proj}}\):
\begin{equation}
    T_{\text{geo}} = W^{3D}_{\text{proj}} F_{\text{geo}} \in \mathbb{R}^{T \times D_{\text{MM}}}.
    \label{eq:proj}
\end{equation}
Finally, the projected geometry tokens are serialized as an independent segment delimited by boundary tokens (\(E_{\text{start\_g}}, E_{\text{end\_g}}\)) and concatenated with text (\(T_{\text{text}}\)) and 2D visual tokens (\(T_{\text{vision}}\)):
\begin{equation}
    H_{\text{input}} = \big[\,T_{\text{text}} \;\oplus\; T_{\text{vision}} \;\oplus\; E_{\text{start\_g}} \;\oplus\; T_{\text{geo}} \;\oplus\; E_{\text{end\_g}}\,\big].
    \label{eq:concat}
\end{equation}

\subsection{Activating Internal Spatial Awareness}
\label{sec:training}

% We adopt a two-stage training strategy to progressively endow the model with spatial capabilities, as illustrated in the overall pipeline of Figure~\ref{fig:pipeline}.

We employ a two-stage training strategy to progressively instill adaptive spatial capabilities, transitioning the model from basic modality alignment to autonomous geometric reasoning, as shown in Figure~\ref{fig:pipeline}.

% \subsubsection{Alignment of 3D Geometric Information}
% \subsubsection{Geometric Feature Alignment}
% \label{sec:alignment}

\noindent\textbf{Geometric Feature Alignment.} 
% The initial phase focuses on projecting independent 3D geometric features into the shared latent space. 
As shown in Fig.~\ref{fig:pipeline}, the initial phase projects 2D and 3D features into a unified embedding space via dual projection modules.
During this stage, the pre-trained feature extractors ($\mathcal{E}_{\text{vision}}$ and $\mathcal{E}_{\text{geo}}$) are frozen to preserve their representational integrity, while the LLM backbone, the 2D/3D projectors ($W_{\text{proj}}^{\text{2D}}$, $W_{\text{proj}}^{\text{3D}}$), and boundary tokens are optimized. This ensures that 3D tokens are semantically aligned with the existing visual-textual space. To provide a robust foundation for this alignment, we utilize a mixture of temporal and spatial grounding datasets, including LLaVA-Hound-64k~\cite{zhang2024direct} and Spar-234k~\cite{wu2025spatial}.

% \subsubsection{Spatial-Aware Supervised Fine-Tuning}
% \label{sec:sft_aware}

% While the alignment phase ensures feature compatibility, it does not guarantee efficient utilization. In the second stage, we aim to evolve the model from a passive receiver into an active perceiver, establishing the ``Internal Sense Decision" mechanism visualized in the top panel of Figure~\ref{fig:pipeline}.

% Instead of indiscriminately processing 3D data, the model is trained to function as a cognitive gate. As depicted by the branching paths in Figure~\ref{fig:pipeline}, the model must dynamically decide whether to integrate geometric information based on the task intent. 
% Specifically, we fine-tune the model using our curated perception dataset (Section~\ref{sec:data}). For tasks necessitating spatial precision, the model learns to generate an explicit trigger token (shown as \texttt{\textless vggt\textgreater} in the "Answer" block of Figure~\ref{fig:pipeline}). This token activates the inclusion of 3D embeddings (the green path); otherwise, the model suppresses the geometry channel to rely solely on 2D context. This mechanism ensures that the computational cost of 3D processing is incurred only when distinctively beneficial.

\noindent\textbf{Spatial-Aware Supervised Fine-Tuning.} Building upon the aligned feature space, this stage evolves the model into an active perceiver capable of making ``Internal Sense Decision''. Rather than treating geometry input as mandatory, the model is trained to function as an adaptive cognitive gate.
As shown by the branching paths in Figure~\ref{fig:pipeline}, the model learns to dynamically evaluate visual input and the task intent. For tasks requiring high spatial precision, the model is trained to trigger a signal, \ie, \texttt{<vggt>}. It acts as a request for geometric embeddings; if the signal is not triggered, the model suppresses the channel to maintain 2D visual reasoning purity. This mechanism ensures that geometric processing is invoked only when necessary, effectively mitigating the noise and computational overhead associated with rigid geometry integration.

\section{Data Curation}
% Encapsulating Empirical Priors}
\label{sec:data}

\begin{figure}[t!]
\vskip 0.1in % 稍微减小顶部间距
\begin{center}
% 修改 width 参数来控制大小 (0.75\columnwidth 比较合适，既不占满也不太小)
\centerline{\includegraphics[width=0.8\columnwidth]{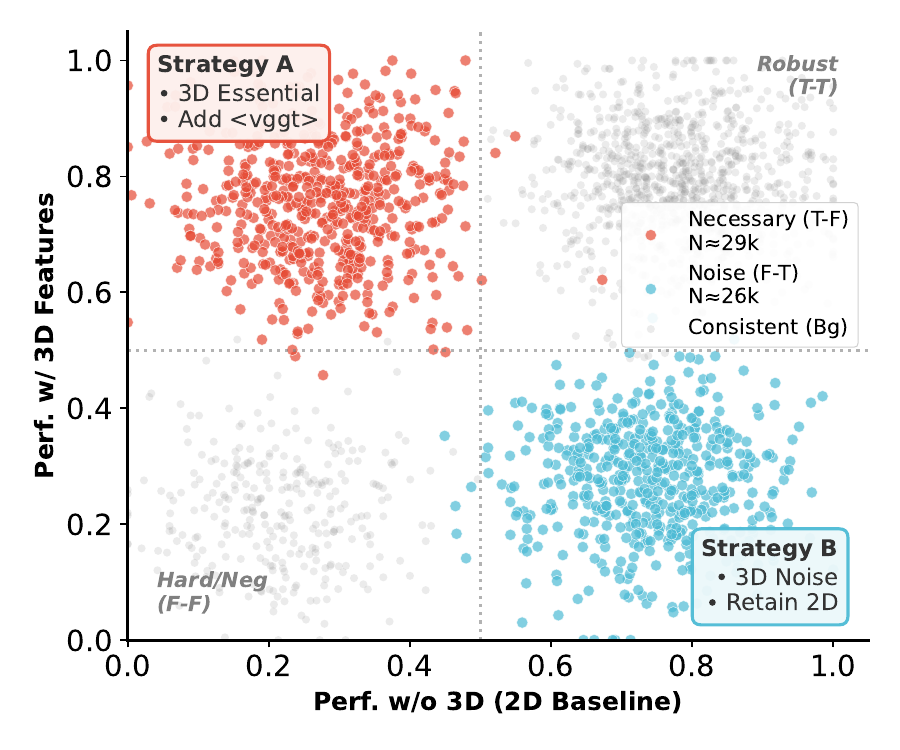}}
\caption{Sample distribution of the perception dataset by training objective. Consistent refers to samples with prediction invariance to 3D features, kept as is. Strategy A and Strategy B represent data where 3D geometric cues are essential and where they should be taken as noise, respectively.}
\label{fig:quad}
\end{center}
\vskip -0.2in
\end{figure}

\begin{table}[t]
\caption{Composition of the Perception Tuning Data.}
% \footnotesize
\label{tab:dataset_stats}

% \vskip 0.15in
\vspace{-1.5mm}

\centering
% 1. 进一步降低行高 (0.9)，极限紧凑
\renewcommand\arraystretch{0.9} 
% 2. 稍微增加一点列间距，平衡缩放后的视觉感受
\setlength{\tabcolsep}{20pt}   
    
    % 3. 将宽度设为 0.85\linewidth (而非全宽)，这能有效"缩小字体"
    % 因为内容较少，铺满全宽会导致字体过大；缩小宽度即缩小字体。
    \resizebox{0.9\linewidth}{!}{%
    \begin{tabular}{l l c}
    \toprule
    
    % --- 表头：深蓝底 + 白字 ---
    % \rowcolor{DeepBlue}
    \textbf{\color{black}Dataset Source} & \textbf{\color{black}Subset / Type} & \textbf{\color{black}Data Size} \\

    % --- 白色分割线 ---
    % \arrayrulecolor{white}
    \midrule
    \arrayrulecolor{black} 

    % --- 数据内容 ---
    
    VSI-590K & Sampled Subset & 55K \\
    \midrule
    
    % SophiaVL-R1 组
    \multirow{3}{*}{SophiaVL-R1} & Chart & 5K \\
     & General & 15K \\
     & Knowledge & 3K \\
    \midrule
    
    % Mantis-Instruct 组
    \multirow{2}{*}{Mantis-Instruct} & LRV-Multi & 13K \\
     & NLVR2 & 16K \\
    \midrule
    
    Llava-Hound-64K & Sampled Subset & 10K \\
    
    \bottomrule
    \end{tabular}
    }
\vskip -0.1in
\end{table}

To extract and internalize the model's intrinsic empirical priors, we construct a comprehensive hybrid dataset derived from VSI-590k~\cite{yang2025cambrian}, SophiaVL-R1-130k~\cite{fan2025sophiavl}, and Mantis-Instruct~\cite{Jiang2024MANTISIM}. We perform dual-condition inference across these samples, generating predictions both with and without 3D geometric features. As illustrated in Fig.~\ref{fig:quad}, samples are categorized into four quadrants based on accuracy: \textit{True-True (T-T)}, \textit{True-False (T-F)}, \textit{False-True (F-T)}, and \textit{False-False (F-F)}.

% Notably, contrary to the conventional assumption that dataset labels align with a single task attribute, our experiments reveal that even within VSI-590K~\cite{yang2025cambrian}—a dataset specifically tailored for indoor spatial reasoning—there exist approximately 25,000 samples exhibiting performance discrepancies (i.e., the union of True-False and False-True cases). Addressing this insight, we implement a differentiated data reconstruction strategy:

Our experiments reveal a critical discrepancy in existing dataset. Even within VSI-590K, which is tailored for indoor spatial reasoning, approximately 25,000 samples exhibit performance divergence when 3D features are introduced. To address this, we implement a differentiated data reconstruction strategy:

% \begin{itemize}
%     \item For True-False samples (where 3D features are a necessary condition), we restructure the data into a two-turn dialogue format. The first turn utilizes a prompt template to generate a targeted Chain-of-Thought (CoT), explicitly appending the control token \textless vggt\textgreater at the end as the concrete instantiation of the model's \textit{Internal Trigger}. The second turn then outputs the original ground truth.
%     \item Conversely, for False-True samples (where 2D features are superior or 3D input introduces noise), we retain the original labels without activating the trigger.
% \end{itemize}

\begin{itemize} 
\item \textbf{Necessary Geometry (Strategy A):} For \textit{T-F} samples where 3D cues are essential, we restructure the data into a two-turn dialogue. The first turn prompts the model to generate a Chain-of-Thought (CoT) that concludes with the \texttt{<vggt>} trigger signal, explicitly stating the need for geometric integration. The second turn then provides the final answer. 
\item \textbf{Geometry as Noise (Strategy B):} For \textit{F-T} samples where 2D features are superior, we retain the original labels while specifically training the model to suppress the trigger signal, thereby relying solely on 2D visual context. \end{itemize}

As summarized in Table~\ref{tab:dataset_stats}, this methodology yields 117k task-aware training samples. This strategy is methodologically significant as it decouples scene context from task types. By ensuring that identical scenes can trigger opposite supervision signals (activation vs. suppression of geometry input), we effectively mitigate shortcut learning. The model is thus compelled to make ``Internal Sense Decisions'' based on a genuine perception of information necessity rather than memorizing background templates.

% As shown in~\cref{tab:dataset_stats}, this methodology yields a total of 117K task-aware training samples. This construction strategy holds critical methodological significance: it decouples the inherent binding between scene context and task type. Consequently, samples derived from the same scene series or involving identical characters may correspond to diametrically opposite supervision signals (i.e., activating vs. suppressing the geometry channel). This mechanism effectively mitigates shortcut learning, where the model might otherwise feign reasoning by memorizing scene backgrounds or specific phrasing templates, thereby compelling the model to make decisions based on the genuine perception of information necessity.
\section{Experiments}
\label{sec:exp}
% In this section, we first describe the implementation details of our model and the experimental setup (Section~\ref{subsec:setup})  including the models involved and the benchmarks adopted. We then present a comprehensive comparison between our GeoSense and the state-of-the-art MLLMs on spatial reasoning and general visual benchmarks (\cref{subsec:performance}). Subsequently, we report the results of ablation studies to demonstrate the effectiveness of each component in our method (\cref{subsec:ablation}). Finally, we conduct confidence-based experiments to qualitatively evaluate the effectiveness of our approach (\cref{subsec:qualitative}).

In this section, we first detail the experimental setup and implementation (\cref{subsec:setup}), followed by a comparison against SOTA MLLMs on spatial reasoning and general visual benchmarks (\cref{subsec:performance}). Finally, we present ablation studies (\cref{subsec:ablation}) and case study of model inference~\cref{subsec:case_study}.

\noindent\textbf{Implementation details.} 
% Our framework is built upon Qwen2.5-VL-3B~\cite{bai2025qwen2}, integrated with VGGT-1B~\cite{wang2025vggt} as the 3D foundation model. In the alignment phase, we fine-tune the model for one epoch on the mixed alignment dataset. We employ the Adam optimizer with a batch size of 32 and a warmup ratio of 0.03. The learning rate is linearly increased to \(1 \times 10^{-6}\) during the warmup phase. To ensure balanced data distribution, each training batch is randomly sampled from a single source within the mixed dataset. For spatial-aware fine-tuning phase, we maintain the same hyperparameter settings as the alignment phase, with the exception of increasing the batch size to 96. Throughout both training stages, as illustrated in~\cref{fig:pipeline}, the visual encoder and the 3D geometry encoder remain frozen to preserve their pre-trained representations; only the LLM backbone and the dedicated projection layers are updated. All experiments were conducted on 6 NVIDIA A100 (80GB) GPUs. The training process required approximately 16 hours for alignment tuning and 26 hours for spatial-aware fine-tuning.
% During the warmup phase, the learning rate was gradually increased to 1e-5 before linearly decaying to 0. In each training step, a batch was randomly sampled from a single source from the mixed dataset. Please refer to the appendix for more details.这里之后不仅要写训练细节，也要写测试细节，比如帧数等
Our framework is built upon the Qwen2.5-VL-3B~\cite{bai2025qwen2}, integrated with VGGT-1B~\cite{wang2025vggt} as the 3D foundation model. We utilize a two-stage training process: \textit{Alignment Phase:} The model is fine-tuned for one epoch on a mixed alignment dataset as stated in Section~\ref{sec:data}, with a batch size of 32. Each training batch is randomly sampled from a single source to ensure balanced data distribution. We use the Adam optimizer~\cite{adam2014method} with a learning rate of $1\times 10^{-6}$ and a 0.03 warmup ratio.
\textit{Spatial-Aware Phase:} We maintain the initial hyperparameters but increase the batch size to 96.
Throughout both stages, the visual and 3D encoders remain frozen, and only the LLM backbone and projection layers are optimized. Experiments were conducted on 8 NVIDIA A100 (80GB) GPUs, requiring 14 hours for alignment and 20 hours for spatial-aware fine-tuning.

% \subsection{Performance Comparison}
% \label{subsec:performance}
\subsection{Experimental Setup}
\label{subsec:setup}

% \subsubsection{Experimental Settings}
To provide a rigorous evaluation of spatial intelligence and general capabilities, we benchmark our approach against baseline models and several representative state-of-the-art models across a comprehensive suite of 10 diverse datasets.
\paragraph{Baseline Models.} We compare against two categories of baselines: \textit{Specialized Spatial Models} and \textit{General Multi-modal Baselines}.
For specialized models, we evaluate SpatialLadder-3B~\cite{li2025spatialladderprogressivetrainingspatial}, which employs a progressive curriculum training with Group Relative Policy Optimization (GRPO), and ViLaSR-7B~\cite{wu2025reinforcingspatialreasoningvisionlanguage}, which integrates an ``Interwoven Thinking and Visual Drawing'' mechanism for embodied reasoning.
In the video domain, SpaceR-sft-7B~\cite{ouyang2025spacerreinforcingmllmsvideo} targets spatiotemporal geometry, while Cambrian-S-3B~\cite{yang2025cambrian} utilizes a surprise-driven memory mechanism for spatial supersensing.
We also include VG-LLM, which injects explicit 3D vision geometry priors, and VST-3B-SFT~\cite{yang2025visualspatialtuning}, a data-centric approach using visual spatial tuning.
For general baselines, we utilize Qwen2.5-VL (7B/3B)~\cite{bai2025qwen2}, featuring Naive Dynamic Resolution and absolute time encoding, and InternVL3-2B~\cite{chen2024internvl}, which demonstrates the efficiency of native multi-modal pre-training with Variable Vision Positional Encoding (V2PE).

\paragraph{Spatial Reasoning Benchmarks.}To rigorously assess spatial intelligence, we employ seven benchmarks across diverse dimensions.
VSI-Bench~\cite{yang2025thinking} is used for holistic 3D understanding, including configurational and measurement tasks.
For multi-view and mental simulation capabilities, we utilize MMSI-Bench~\cite{yang2025mmsibenchbenchmarkmultiimagespatial} and MindCube-Tiny~\cite{yin2025spatialmentalmodelinglimited}, which require constructing mental models from limited views.
Low-level geometric perception is evaluated using BLINK~\cite{fu2024blinkmultimodallargelanguage}, which resists language mediation, and the 3D subset of CV-Bench~\cite{tong2024cambrian} for depth and occlusion reasoning.
Additionally, EmbSpatial~\cite{du2024embspatialbenchbenchmarkingspatialunderstanding} is employed to test egocentric spatial understanding for embodied tasks.
We report the arithmetic mean of these datasets as the \textit{Spatial Avg.}.

\paragraph{General Visual Benchmarks.}To ensure that spatial specialization does not compromise general capabilities, we evaluate models on four established benchmarks.
MMBench~\cite{liu2024mmbenchmultimodalmodelallaround} provides a comprehensive assessment using CircularEval to mitigate bias.
We use MME~\cite{fu2025mmecomprehensiveevaluationbenchmark} (Perception Score) to measure recognition breadth and POPE~\cite{li2023evaluatingobjecthallucinationlarge} to strictly detect object hallucinations.
Furthermore, WeMath~\cite{qiao2024wemathdoeslargemultimodal} is included as a proxy for high-level logical reasoning and geometric abstraction.
The \textit{General Avg.} aggregates these scores to reflect overall model robustness.

\subsection{Performance Comparison}
\label{subsec:performance}

\begin{table*}[t]
% 标题更新：明确指出按照 Data Size 排序
\caption{Performance comparison on Spatial and General benchmarks. We report average scores for each benchmark collection, seperate benchmarks, and a total average for ranking. For spatial reasoning benchmarks, we include VSI-Bench~\cite{yang2025thinking}, MMSI~\cite{yang2025mmsibenchbenchmarkmultiimagespatial}, MindCube (abbr. as MC)~\cite{yin2025spatialmentalmodelinglimited}, BLINK~\cite{fu2024blinkmultimodallargelanguage}, EmbSpatial~\cite{du2024embspatialbenchbenchmarkingspatialunderstanding} and CV-Bench(abbr. as CVB)~\cite{tong2024cambrian} for comprehensive spatial reasoning evaluation. For general benchmarks, we include MMBench~\cite{liu2024mmbenchmultimodalmodelallaround}, MME~\cite{fu2025mmecomprehensiveevaluationbenchmark}, POPE~\cite{li2023evaluatingobjecthallucinationlarge}, and WeMath~\cite{qiao2024wemathdoeslargemultimodal} to evaluate the general multimodal reasoning capability. Note that we evaluate the 3D subset of CV-Bench and the MME perception score. * denotes fine-tuning on the same data as our model used.}
\label{tab:main_results}

% \vskip 0.15in
% \vspace{-1mm}

\centering
\renewcommand\arraystretch{1.1} 
\setlength{\tabcolsep}{1.2mm}{ 
    \resizebox{1.0\linewidth}{!}{%
    \begin{tabular}{l | c | ccccccc | ccccc | cc}
    \toprule

    % --- 第一层表头 ---
    \multicolumn{2}{c}{} & 
    \multicolumn{7}{c}{\cellcolor{teal!10}\textbf{Spatial Reasoning Benchmarks}} & 
    \multicolumn{5}{c}{\cellcolor{yellow!10}\textbf{General Benchmarks}} & 
    \multicolumn{2}{c}{\textbf{Overall}} \\ 

    \cmidrule(lr){1-9} \cmidrule(lr){10-14} \cmidrule(lr){15-16}

    % --- 第二层表头 ---
    \textbf{Model} & \textbf{FT-Data} & 
    % Spatial
    \rotatebox{45}{\textbf{\textit{Avg.}}} & 
    \rotatebox{45}{\textbf{VSI-Bench}} & \rotatebox{45}{\textbf{MMSI}} & \rotatebox{45}{\textbf{MC-Tiny}} & \rotatebox{45}{\textbf{BLINK}} & \rotatebox{45}{\textbf{EmbSpatial}} & \rotatebox{45}{\textbf{CVB(3D)}} & 
    % General
    \rotatebox{45}{\textbf{\textit{Avg.}}} & 
    \rotatebox{45}{\textbf{MMBench}} & \rotatebox{45}{\textbf{MME (P)}} & \rotatebox{45}{\textbf{POPE}} & \rotatebox{45}{\textbf{WeMath}} & 
    % Overall
    \rotatebox{45}{\textbf{Avg.}} & \rotatebox{45}{\textbf{Rank}} \\ 
    \midrule

    % --- 数据区域 (Baseline 按数据量从小到大排序) ---
    Qwen2.5-VL-3B~\cite{bai2025qwen2} & - & 43.4 & 27.0 & 28.6 & 37.6 & 33.1 & 62.3 & 71.8 & 53.3 & 76.6 & 1526.1 & 87.5 & 25.8 & 48.3 & 11 \\
    Qwen2.5-VL-7B~\cite{bai2025qwen2} & - & 50.5 & 32.3 & 26.8 & 36.0 & 55.9 & 71.8 & 80.1 & 57.8 & 82.6 & 1693.9 & 87.8 & 33.8 & 54.1 & 3 \\
    InternVL3-2B~\cite{chen2024internvl} & - & 43.6 & 33.0 & 26.5 & 37.5 & 30.0 & 60.1 & 74.3 & 54.1 & 78.6 & 1610.2 & 88.9 & 22.4 & 48.8 & 9 \\
    % qwen2.5-vl-7B (Instruct) & $\sim$4t & 50.5 & 32.3 & 26.8 & 36.0 & 55.9 & 71.8 & 80.1 & 57.8 & 82.6 & 1693.9 & 87.8 & 33.8 & 54.1 & 4 \\
    \midrule
    ViLASR-7B~\cite{wu2025reinforcingspatialreasoningvisionlanguage} & {\small 73K} & \underline{51.0} & 44.6 & 30.2 & 35.1 & 51.4 & 67.3 & 77.2 & 54.5 & 80.8 & 1634.4 & 84.8 & 25.3 & 52.8 & 4 \\
    SpaceR-sft-7B~\cite{ouyang2025spacerreinforcingmllmsvideo} & {\small 151K} & 49.9 & 41.6 & 27.4 & 38.0 & 49.6 & 66.9 & 75.7 & 58.9 & 82.8 & 1688.1 & 88.0 & 39.0 & 54.4 & 2 \\
    % InternVL3-2B & 21m & 43.6 & 33.0 & 26.5 & 37.5 & 30.0 & 60.1 & 74.3 & 54.1 & 78.6 & 1610.2 & 88.9 & 22.4 & 48.8 & 10 \\
    % \midrule
    
    SpatialLadder-3B~\cite{li2025spatialladderprogressivetrainingspatial} & {\small 26K} & 48.5 & 44.9 & 27.4 & 43.5 & 43.0 & 58.2 & 73.7 & 52.5 & 72.3 & 1403.2 & 85.5 & 34.4 & 50.5 & 7 \\
    Cambrain-S-3B~\cite{yang2025cambrian} & {\small 10M} & 49.6 & 56.1 & 27.0 & 38.4 & 37.7 & 63.5 & 75.2 & \textbf{55.4} & 74.9 & 1490.8 & 86.8 & 40.7 & 52.5 & 5 \\
    % VST-3B-SFT~\cite{yang2025visualspatialtuning} & {\small 4.2M} & 54.8 & 51.4 & 28.8 & 36.0 & 58.8 & 69.0 & 84.9 & 56.6 & 79.6 & 1458.7 & 87.8 & 42.9 & 55.7 & 2 \\
    VG-LLM~\cite{zheng2025learning} & {\small 385k} & 49.7 & 47.3 & 27.6 & 31.2 & 48.9 & 64.2 & 79.1 & 52.0 & 75.4 & 1441.6 & 86.9 & 25.6 & 50.9 & 6 \\

    \midrule
    VG-LLM$^{\bf *}$~\cite{zheng2025learning} & {\footnotesize 940k} & 48.8 & 52.3 & 25.2 & 30.1 & 62.8 & 50.9 & 71.7 & 48.2 & 73.2 & 1248.5 & 74.2 & 31.1 & 48.5 & 10 \\
    Qwen2.5-VL-3B$^{\bf *}$~\cite{bai2025qwen2} & {\footnotesize 940k} & 49.2 & 48.1 & 24.2 & 41.5 & 63.1 & 50.4 & 67.9 & 51.0 & 71.0 & 1301.2 & 85.2 & 33.8 & 50.1 & 8 \\
    
    % Ours 放在最后
    \rowcolor{blue!5} \textbf{GeoSense} & {\small 940K} & \textbf{\textit{56.6}} & 54.9 & 27.5 & 45.1 & 68.5 & 64.3 & 78.7 & \textit{\underline{55.2}} & 75.9 & 1473.7 & 85.4 & 41.2 & \textbf{55.9} & 1 \\
    \bottomrule
    \end{tabular}
    }
}
\end{table*}

\begin{table*}[t]
\caption{Ablation study on different geometry injection strategy. We compare the baseline against the additive fusion manner (VG-LLM) and our adaptive independent token insertion approach (GeoSense). All variants are trained with the same data to ensure fair comparison.}
\label{tab:vggt_ablation}

% \vskip 0.15in

\centering
\renewcommand\arraystretch{1.15} 
\setlength{\tabcolsep}{1.2mm}    
    \resizebox{1.0\linewidth}{!}{%
    % 【修改点1】在第1列(l)和第2列(c)之间也添加竖线 |，变成 l | c | ...
    \begin{tabular}{l | c | cccccccc cccc} 
    \toprule
    
    % --- 第一行：大标题分组 ---
    % 【修改点2】将 \multicolumn{2}{c|}{} 拆分为两个 \multicolumn{1}{c|}{}
    % 这样第一行的空白处中间也会有一条竖线
    \multicolumn{1}{c|}{} & \multicolumn{1}{c|}{} & 
    % 分组 1: VSI-Bench
    \multicolumn{8}{c}{\cellcolor{teal!10}\textbf{VSI-Bench}} & 
    % 分组 2: MindCube
    \multicolumn{4}{c}{\cellcolor{yellow!10}\textbf{MindCube-Tiny}} \\
    
    % 分组划线
    \cmidrule(lr){3-10} \cmidrule(lr){11-14}
    
    % --- 第二行：具体列名 ---
    % 继承上方的竖线设置
    \textbf{Model} & \textbf{Injection} & 
    % VSI Columns
    \textbf{Avg.} & \textbf{Obj. Count} & \textbf{Obj. Size} & \textbf{Room Size} & \textbf{Rel. Dir.} & \textbf{Route Plan} & \textbf{Appr. Order} & \textbf{Rel. Dist.} & 
    % MindCube Columns
    \textbf{Avg.} & \textbf{Rotation} & \textbf{Among} & \textbf{Around} \\
    \midrule

    % --- 数据行 ---
    
    Qwen2.5-vl-3B & 0.00\% & 48.07 & 63.45 & 58.45 & 41.94 & 46.02 & 31.95 & 59.87 & 49.43 & 35.67 & \textbf{34.50} & 41.50 & 27.50 \\
    
    + fusion VGGT (VG-LLM) & 100.00\% & 52.34 & 64.67 & 63.96 & 39.06 & \textbf{61.61} & \textbf{35.56} & 61.00 & \textbf{54.78} & 30.58 & 34.00 & 30.00 & 29.75 \\
    
    % 最后一行高亮
    \rowcolor{blue!5} + alternative VGGT (GeoSense) & 35.68\% & \textbf{54.86} & \textbf{69.43} & \textbf{66.64} & \textbf{58.26} & 52.85 & 34.02 & \textbf{71.52} & 53.66 & \textbf{45.08} & 31.50 & \textbf{49.17} & \textbf{45.75} \\
    
    \bottomrule
    \end{tabular}
    }
\end{table*}

As demonstrated in~\cref{tab:main_results}, our proposed model significantly outperforms the baseline Qwen2.5-VL-3B \cite{bai2025qwen2} on spatial reasoning benchmarks under comparable training data scales. Notably, our model achieves competitive spatial reasoning performance relative to the substantially larger Qwen2.5-VL-7B \cite{bai2025qwen2}. Concurrently, our model performs well for general visual reasoning tasks by dynamically reverting to standard 2D and textual embedding combinations. The system effectively mitigates cross-modal interference typically caused by redundant geometric features in non-spatial contexts.

\noindent\textbf{Performance on spatial reasoning.} Specifically, our model achieves superior comprehensive performance by internalizing the perception of spatial feature requirements. For instance, in the BLINK~\cite{fu2024blinkmultimodallargelanguage} benchmark, which comprises diverse spatial subtasks with varying dependencies on geometric relations, our model attains the highest overall score. This reinforces our objective: enabling the model to adaptively utilize geometric features based on task demands, thus enhancing spatial reasoning while preserving general inference capabilities. Furthermore, to isolate the impact of training data, we compare our model with the fine-tuned baseline (marked with {\footnotesize $^{*}$}) using identical datasets. On the EmbSpatial~\cite{du2024embspatialbenchbenchmarkingspatialunderstanding} subset, while the VG-LLM model (which consistently introduces 3D features) exhibits a performance decline after fine-tuning, our approach effectively improves spatial reasoning through perception-driven sample selection. This underscores that our gains derive from the efficacy of the mechanism rather than merely data quality.

\noindent\textbf{Performance on general visual reasoning.} For general visual tasks, Our model tends to suppress redundant interference, with geometric features being triggered for only approximately 3\% of samples. Surprisingly, our model also achieves superior performance on the WeMath~\cite{qiao2024wemathdoeslargemultimodal} dataset, which could be attributed to a significant portion of the test set involving spatial-geometric mathematical problems. Compared to models that blindly incorporate geometric features, our approach yields a significant improvement, demonstrating that an activated perceptual capability allows the model to retain essential spatial imagination for complex problem-solving (\eg, math and physics) even without explicit external geometric inputs.

\subsection{Ablation Study}
\label{subsec:ablation}
% \subsubsection{Comparison of VGGT Used Scheme}

\noindent\textbf{Comparison of geometry injection scheme.} We conducted an ablation study using identical training data to evaluate three VGGT~\cite{wang2025vggt} integration schemes: (1) w/o VGGT (baseline), (2) Visual Fusion (element-wise addition), and (3) our proposed Adaptive Selection. Detailed comparisons were performed on two mainstream spatial reasoning benchmarks: VSI-Bench~\cite{yang2025thinking} and MindCube-Tiny~\cite{yin2025spatialmentalmodelinglimited}. Experimental results demonstrate that our adaptive scheme activated VGGT features as auxiliary input tokens for an average of 35.68\% of samples. Notably, the activation rate is 43.7\% on VSI-Bench and 27.58\% on MindCube-Tiny. This variance underscores the flexibility of our approach. For instance, in the Among category of MindCube, the task requires spatial logic (locating cues based on descriptions) rather than spatial reconstruction. Our model correctly identified that geometric features was unnecessary for these samples, thereby avoiding redundant computation.

Specifically, our model outperformed the other two schemes on the majority of VSI-Bench~\cite{yang2025thinking} tasks. A notable exception is the Relative Direction task, where our model lagged behind the Visual Fusion baseline. We attribute this to VGGT feature for multi-scene contexts. Treating VGGT features for dense scene as an independent modality token appears to amplify the signal-to-noise ratio disadvantage compared to feature-level fusion. Nevertheless, the performance gain over the ``w/o VGGT'' baseline confirms the validity of incorporating geometric priors.

In contrast, results on MindCube revealed distinct behavior. For the Rotation task, all schemes utilizing VGGT exhibited performance degradation. This is likely due to the data characteristics (wide-angle views with only 2-4 images), which hinder the effective construction of global representations by VGGT. Interestingly, the Visual Fusion scheme performed slightly better than ours in this specific failure case. We hypothesize that fusing VGGT with strong 2D features implicitly dilutes the negative impact of the low-quality geometric representation, whereas our approach exposes the model more directly to the suboptimal signals.

\begin{table*}[t!]
    % \vspace{-2.5mm}
    \centering
    % --- 左侧：表格 Minipage (占用 68% 宽度) ---
    % [c] 表示垂直居中对齐，如果想顶部对齐改用 [t]
    \begin{minipage}[c]{0.68\linewidth}
        \captionof{table}{Ablation Study on baseline and our proposed variants (GeoSense-4B) with different training stages. MME scores are scaled by 28 for average calculation. \textbf{Bold} denotes the best performance. \underline{Underline} denotes the second-highest performance.}
        \label{table:ablation_study}
        
        \centering
        \renewcommand\arraystretch{1.15}
        \setlength{\tabcolsep}{1.2mm}
        % 注意：这里的 resizebox 设为 1.0\linewidth，是指占满这个 minipage 的宽度
        \resizebox{1.0\linewidth}{!}{%
            \footnotesize
            % 你的原始表格代码
            \begin{tabular}{c|c|c|cccc|ccc|c}
            \toprule
            \multirow{2}{*}{\#} & \multirow{2}{*}{\textbf{Data}} & \multirow{2}{*}{\textbf{Model}} & 
            \multicolumn{4}{c|}{\textbf{Spatial Tasks}} & 
            \multicolumn{3}{c|}{\textbf{General Tasks}} & 
            \multirow{2}{*}{\textbf{Avg}} \\ 

            \cmidrule(lr){4-7} \cmidrule(lr){8-10} 

            & & & \textbf{VSI-Bench} & \textbf{MindCube} & \textbf{BLINK} & \textbf{CV-Bench} & \textbf{MM-Bench} & \textbf{MME} & \textbf{MM-Star} & \\ 
            \midrule

            1 & $\emptyset$ & Qwen2.5-VL-3B & 
            27.0 & 37.6 & 33.1 & 71.8 & 
            \textbf{76.6} & \textbf{2104} & \textbf{56.2} & 
            51.0 \\ 
            \midrule

            2 & Initial & GeoSense-4B & 
            42.8 & 37.8 & 58.9 & 74.9 & 
            75.2 & 1974 & 51.6 & 
            58.8 \\ 
            
            3 & Align. SFT & GeoSense-4B & 
            \underline{53.9} & \underline{44.4} & \underline{61.9} & \underline{77.7} & 
            72.2 & 1803 & 45.5 & 
            \underline{60.0} \\ 

            \rowcolor{blue!5} 
            4 & Percept. SFT & GeoSense-4B & 
            \textbf{54.9} & \textbf{45.7} & \textbf{68.5} & \textbf{78.7} & 
            \underline{75.9} & \underline{1922} & \underline{51.7} & 
            \textbf{63.4} \\ 

            \bottomrule
            \end{tabular}
        }
    \end{minipage}
    \hfill % 在左右两个 minipage 之间填充空白，使其撑满行宽
    % --- 右侧：图片 Minipage (占用 30% 宽度) ---
    \begin{minipage}[c]{0.30\linewidth}
        \centering
        % 插入你的图片，确保路径正确
        % 这里的 width=\linewidth 指的是占满这 30% 的区域
        \includegraphics[width=\linewidth]{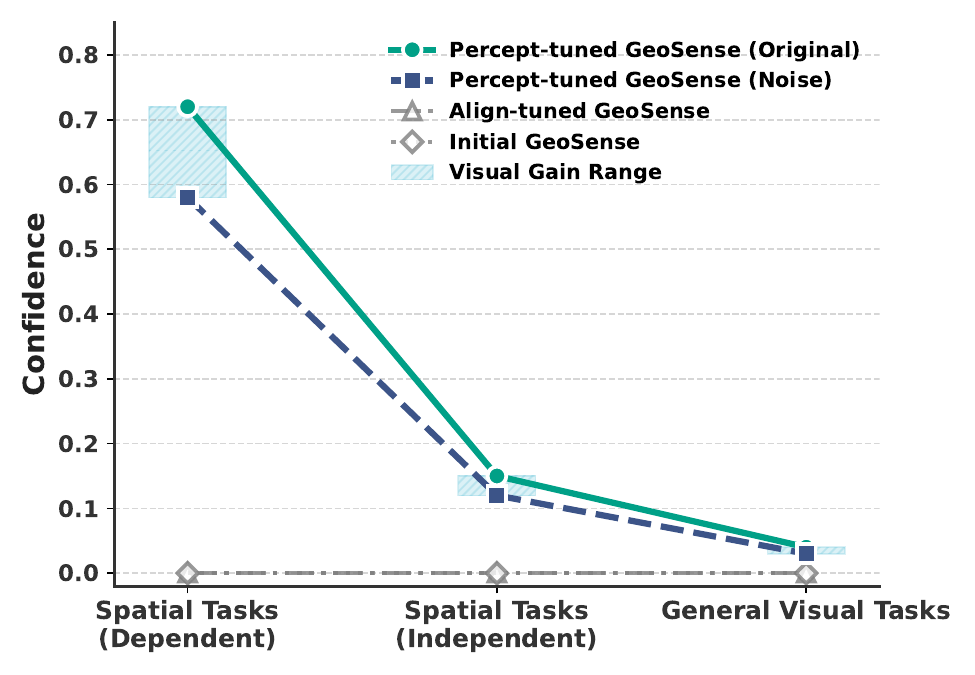} 
        
        \vspace{-5pt} % 调整图片和 Caption 的间距
        
        % 关键点：在 table 环境里给图片加标题，必须用 \captionof{figure}
        \captionof{figure}{Confidence scores of 3D trigger token across spatial and general tasks.}
        \label{fig:confidence}
    \end{minipage}
    
    % \vskip -0.1in % 调整底部间距
\end{table*}

\begin{figure*}[t!] % 关键点1：加星号 *；关键点2：通常使用 [t] 置顶
    % \vskip -0.1in % ICML 模版建议的顶部间距
    % \vspace{-2mm}
    \begin{center}
        % 关键点3：宽度通常设为 \linewidth 或 \textwidth (占满通栏)
        % 如果觉得太大，可以改为 0.8\linewidth
        \centerline{\includegraphics[width=1\linewidth]{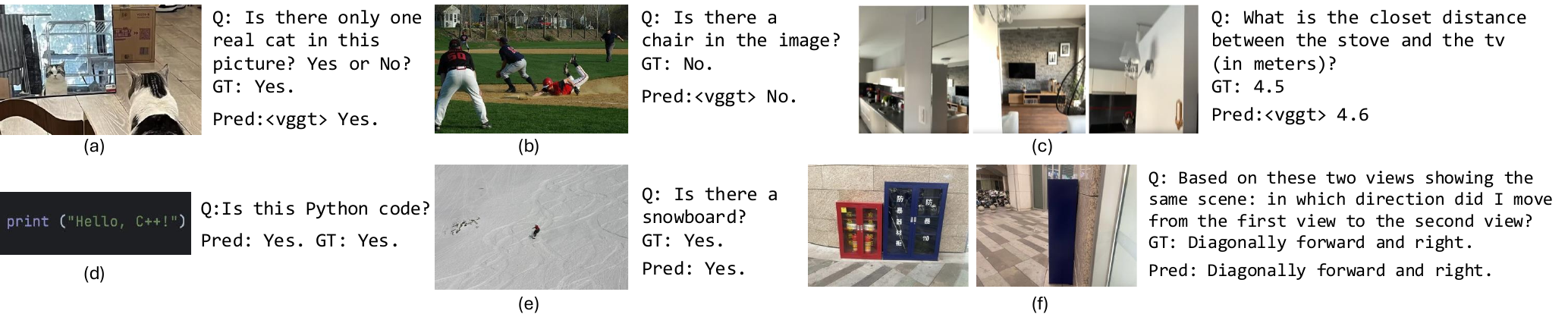}}
        \vspace{-2mm}
        
        \caption{\textbf{Case study of internal sense decision.} We present representative examples demonstrating how our model adaptively determines whether to trigger 3D geometric features according to the input and task. \textbf{(a, b)}: Rare cases where general tasks explicitly demand geometric embedding. \textbf{(c)}: Typical activation for spatial reasoning. \textbf{(d, e)}: Standard suppression for general visual inputs. \textbf{(f)}: Spatial queries solved effectively without geometric triggers.}
        \label{fig:case}
    \end{center}
    % \vskip -0.2in % ICML 模版建议的底部间距修正
    \vspace{-5mm}
\end{figure*}

\noindent\textbf{Effectiveness of components.} To rigorously validate the efficacy of each component within our proposed framework, we conducted a comprehensive ablation study comparing the model's performance trajectories across different training stages.
Specifically, we evaluated three variations:
(1) The Initialized Model, which integrates the pre-trained weights of Qwen2.5-VL-3B~\cite{bai2025qwen2} and VGGT-1B~\cite{wang2025vggt}, while notably initializing the projection layer with weights from a state-of-the-art model to map geometric features into the 2D visual embedding space;
(2) The Alignment-Tuned Model;
and (3) The final Perception-Tuned Model.
We benchmarked these variants against the vanilla Qwen2.5-VL-3B baseline across both spatial and general visual tasks.

\noindent\textbf{Confidence score of trigger token under difference setting.} As shown in the~\cref{table:ablation_study}, spatial reasoning tasks exhibit a consistent monotonic improvement, with the average score surging from 51.0 to 63.4. Notably, on the challenging BLINK benchmark, our method achieves a remarkable gain (+35.4) compared to the baseline. Conversely, general visual reasoning tasks display a ``dip-then-recover'' trajectory. We attribute the Initial model's relative robustness to the pre-trained projection layer, which maps geometric embeddings into a pseudo-2D visual space, effectively treating VGGT inputs as implicit visual features. However, the subsequent introduction of special tokens during Align. SFT disrupts this implicit mapping, causing a temporary performance degradation (\eg, MME drops to 1803). Crucially, the proposed Percept. SFT reverses this decline. By learning to selectively gate VGGT embeddings, the model mitigates interference, restoring general capabilities (\eg, MME recovers to 1922) while maintaining peak spatial performance. This confirms our design objective: significantly enhancing spatial reasoning while preserving the integrity of general multimodal understanding.

Similarly, we analyzed the confidence contributions of different components regarding the trigger token. As illustrated in~\cref{fig:confidence}, we examined the probabilistic behavior of this token across three distinct task categories: geometry-reliant tasks, geometry-agnostic spatial tasks, and general visual tasks. Our results indicate that high confidence scores are exhibited exclusively by the model following perceptual fine-tuning, whereas previous training stages fail to facilitate the triggering of such perceptual signals. To verify that this decision is not driven solely by textual priors, we conducted a control experiment by replacing the input image with random noise. Results demonstrate a significant attenuation in the confidence of the spatial query token under noise conditions. This finding confirms that the model's determination of spatial information requirements is contingent upon a multi-modal synergy between textual prompts and underlying visual semantics, underscoring the effectiveness of our approach in achieving robust, context-dependent feature modulation.
% Our empirical results reveal profound insights into the complexity of integrating geometric priors.
% Despite equipping the Initialized Model with a pre-aligned projector, we observed that directly incorporating these geometric embeddings presents a significant distributional challenge. While this initialization yields a relative improvement in spatial reasoning compared to the baseline, it concurrently causes a marked degradation in general visual capabilities. This phenomenon underscores that a structural connection via a projector alone is insufficient; our proposed training paradigm is essential to bridge the semantic gap effectively.

% Furthermore, during the alignment phase—where we explicitly enforce the utilization of VGGT features—the model exhibits a sharp increase in spatial understanding. However, this comes at the cost of general capability (reflecting a ``sawtooth'' performance trade-off), confirming that the model is actively adapting to the geometric modality.
% Crucially, the subsequent perception tuning stage enables the model to contextually discern task requirements. This phase not only further refines spatial reasoning but effectively restores general visual proficiency to a high level.
% Collectively, the stepwise performance improvements corroborate that each component plays an indispensable role in achieving robust, generalized spatial intelligence.

\subsection{Capability on Embodied Task}
To further examine how our model performs in more realistic embodied settings, we evaluate it on the recent Quantiphy-validation benchmark showed in~\cref{tab:embodied}. Compared with both general-purpose and specialised models of a similar scale, our model achieves the best overall performance. Owing to the relatively balanced composition of 2D and 3D tasks in this benchmark, the results suggest that our approach is able to maintain a reasonable trade-off between 2D perception and 3D geometric reasoning, which is consistent with our earlier analysis.

Although our model does not reach the absolute state-of-the-art in all four sub-scenarios, it consistently ranks within the top three for each individual setting. In particular, it attains the second-best results on both 2D and 3D static tasks, which aligns well with our expectations. We also observe that performance in dynamic scenarios is generally lower than in static ones. We conjecture that this gap mainly stems from the distribution bias in our fine-tuning data, which contains fewer samples involving significant motion and viewpoint variation.

Moreover, apart from our method, none of the baseline models can maintain stable rankings across all four categories. This phenomenon suggests that existing training paradigms for multimodal models still struggle to cope with the feature variations introduced by viewpoint changes and object motion in embodied environments. The absence of a clear overall performance breakthrough further highlights the necessity of constructing training data and optimisation strategies that are more specifically designed for embodied tasks.

% \begin{table}[t]
% \caption{Comparison of different models on embodied task. \textbf{Overall} denotes the aggregate score, while S2/D2 represent static/dynamic 2D sub-task and S3/D3 represent static/dynamic 3D sub-task (reported in \%).}
% \label{tab:embodied}
% \centering
% \resizebox{\linewidth}{!}{% 自动缩放以适应单栏宽度
% \begin{tabular}{lccccc}
% \toprule
% \textbf{Model} & \textbf{Overall} & \textbf{S2} & \textbf{D2} & \textbf{S3} & \textbf{D3} \\
% \midrule
% InternVL3-2B           & 29.25 & 21.88 & 27.30 & 25.58 & 39.15 \\
% Qwen2.5-VL-3B-Instruct & 36.79 & 37.81 & 44.32 & 27.91 & 38.30 \\
% Qwen3-VL-2B-Instruct   & 30.44 & 19.69 & 31.62 & 23.02 & 43.62 \\
% Qwen3-VL-4B-Instruct   & 36.60 & 26.56 & 32.70 & 29.77 & 52.77 \\
% \midrule
% SpatialLadder-3B       & \textit{36.86} & 34.06 & 34.59 & 33.72 & 43.40 \\
% VG-LLM-4B              & 31.13 & 34.06 & 36.76 & 17.67 & 37.02 \\
% VST-3B-SFT             & 33.77 & 22.50 & 29.19 & 29.77 & 48.72 \\
% \midrule
% \textbf{GeoSense-4B (Ours)} & \textbf{37.36} & 36.56 & 35.41 & 31.39 & 44.89 \\
% \bottomrule
% \end{tabular}%
% }
% \end{table}
\begin{table}[t]
\caption{Comparison of different models on embodied task. \textbf{Overall} denotes the aggregate score, while S2/D2 represent static/dynamic 2D sub-task and S3/D3 represent static/dynamic 3D sub-task (reported in \%). The best results are highlighted in \textbf{bold}, and the second-best results are \underline{underlined}.}
\label{tab:embodied}
\centering
\resizebox{\linewidth}{!}{% 自动缩放以适应单栏宽度
\begin{tabular}{lccccc}
\toprule
\textbf{Model} & \textbf{Overall} & \textbf{S2} & \textbf{D2} & \textbf{S3} & \textbf{D3} \\
\midrule
InternVL3-2B           & 29.25 & 21.88 & 27.30 & 25.58 & 39.15 \\
Qwen2.5-VL-3B-Instruct & 36.79 & \textbf{37.81} & \textbf{44.32} & 27.91 & 38.30 \\
Qwen3-VL-2B-Instruct   & 30.44 & 19.69 & 31.62 & 23.02 & 43.62 \\
Qwen3-VL-4B-Instruct   & 36.60 & 26.56 & 32.70 & 29.77 & \textbf{52.77} \\
\midrule
SpatialLadder-3B       & \underline{36.86} & 34.06 & 34.59 & \textbf{33.72} & 43.40 \\
VG-LLM-4B              & 31.13 & 34.06 & \underline{36.76} & 17.67 & 37.02 \\
VST-3B-SFT             & 33.77 & 22.50 & 29.19 & 29.77 & \underline{48.72} \\
\midrule
\textbf{GeoSense-4B (Ours)} & \textbf{37.36} & \underline{36.56} & 35.41 & \underline{31.39} & 44.89 \\
\bottomrule
\end{tabular}%
}
\end{table}

\subsection{Case Study}
\label{subsec:case_study}
% We conducted an analysis of the model's outputs, with representative cases visualized in~\cref{fig:case}. 
We visualize representative model outputs in \cref{fig:case} to analyze the internalized perception of geometric necessity.
Beyond the typical samples annotated for 3D reasoning in the training set, such as distance analysis and object dimension estimation, a particularly compelling result emerged in the query ``How many cats are in the image?'' (~\cref{fig:case} (a)). Here, the model explicitly requested geometric auxiliary features to effectively distinguish between the actual kitten and its mirror reflection, indicating that it has learned to leverage 3D geometric features to discern the spatial attributes of objects. Furthermore, the model utilized geometric cues to filter out interference from occlusion, successfully avoiding the misclassification of the object as a chair (~\cref{fig:case} (b)).

Regarding general visual tasks, the model predominantly learned to suppress geometric feature activation. Notably, in specific simple directional queries that ostensibly appear to require geometric insight, the model derived the correct answer via semantic and commonsense reasoning instead. For instance, in~\cref{fig:case} (f), the spatial relationship could be inferred solely through color correspondence, rendering explicit geometric input unnecessary. Collectively, these examples corroborate that the model has acquired the capability to adaptively perceive the necessity of spatial information based on specific task demands and available context.

\section{Conclusion}
% \vspace{-1mm}
\label{sec:Conclusion}

% In this paper, we introduced \textit{GeoSense}, a framework enabling MLLMs to autonomously perceive geometric necessity. Central to our contribution is the Task-Necessity Perception data curation and framework training pipeline, which extracts the model's intrinsic empirical priors to guide decision-making, and internalize the spatial sense necessity awareness. Empowered by this data-centric approach, GeoSense achieves a state-of-the-art balance: it significantly enhances spatial reasoning capabilities (\eg, on VSI-Bench) while strictly preserving general multimodal performance.

In this paper, we introduced \textit{GeoSense}, a framework that enables MLLMs to autonomously perceive and act upon geometric necessity. Our core contribution is a model-adaptive training pipeline that internalizes spatial awareness by extracting empirical priors directly from the model's own inference discrepancies. 
It treats geometric features as an on-demand resource, invoking them only when standard 2D perception is insufficient. 
Empowered by this data-centric approach, \textit{GeoSense} achieves a superior balance: it establishes new SOTA performance on spatial reasoning benchmarks, while strictly preserving general multimodal reasoning capabilities.

% Experimental results demonstrate that this data-centric approach achieves a superior balance: it establishes new state-of-the-art performance on spatial reasoning benchmarks like VSI-Bench while strictly preserving general multimodal reasoning capabilities. By bridging the perception-task information gap, GeoSense provides a robust and efficient blueprint for deploying spatially-aware MLLMs in complex, real-world environments.

\vspace{-2mm}
\paragraph{Limitations and Future Work.}Currently, our geometric features primarily rely on the VGGT encoder. Future research will explore integrating diverse 3D representations (\eg, point clouds) and further optimizing the internal trigger mechanism to maximize efficiency for deployment on resource-constrained edge devices.

\bibliography{arxiv}
\bibliographystyle{icml2026}

%%%%%%%%%%%%%%%%%%%%%%%%%%%%%%%%%%%%%%%%%%%%%%%%%%%%%%%%%%%%%%%%%%%%%%%%%%%%%%%
%%%%%%%%%%%%%%%%%%%%%%%%%%%%%%%%%%%%%%%%%%%%%%%%%%%%%%%%%%%%%%%%%%%%%%%%%%%%%%%
% APPENDIX
%%%%%%%%%%%%%%%%%%%%%%%%%%%%%%%%%%%%%%%%%%%%%%%%%%%%%%%%%%%%%%%%%%%%%%%%%%%%%%%
%%%%%%%%%%%%%%%%%%%%%%%%%%%%%%%%%%%%%%%%%%%%%%%%%%%%%%%%%%%%%%%%%%%%%%%%%%%%%%%
% \newpage
% \appendix
% \onecolumn

\end{document}